\pdfoutput=1

\documentclass[11pt]{article}

\usepackage[preprint]{acl}
\usepackage{times}
\usepackage{latexsym}

\usepackage[T1]{fontenc}

\usepackage[utf8]{inputenc}

\usepackage{microtype}

\usepackage{inconsolata}

\usepackage{graphicx}
\usepackage{hyperref}
\usepackage{array}

\usepackage{tabularx}
\usepackage{subfigure,graphicx}
\usepackage{booktabs}
\usepackage{multirow}
\usepackage{makecell}
\usepackage{xcolor}
\usepackage{eqparbox} 
\usepackage{transparent}

\newcommand{\ie}{\textit{i.e.}}
\newcommand{\eg}{\textit{e.g.}}

\newcommand{\ours}{{DomainRAG}}

\title{\ours{}: A Chinese Benchmark for Evaluating Domain-specific Retrieval-Augmented Generation}

\author{Shuting Wang$^1$, Jiongnan Liu$^1$, Shiren Song$^1$, Jiehan Cheng$^1$, Yuqi Fu$^1$ \\ \textbf{ Peidong Guo$^2$, Kun Fang$^2$, Yutao Zhu$^1$, }\and \textbf{Zhicheng Dou$^{1*}$} \\
$^1$Gaoling School of Artificial Intelligence, Renmin University of China \\
$^2$Baichuan Intelligent Technology\\
\texttt{\{wangshuting, liujn, dou\}@ruc.edu.cn}
}

\begin{document}
\maketitle

\def\thefootnote{*}\footnotetext{Corresponding author.}
\def\thefootnote{\arabic{footnote}}

\begin{abstract}
Retrieval-Augmented Generation (RAG) offers a promising solution to address various limitations of Large Language Models (LLMs), such as hallucination and difficulties in keeping up with real-time updates. This approach is particularly critical in expert and domain-specific applications where LLMs struggle to cover expert knowledge. Therefore, evaluating RAG models in such scenarios is crucial, yet current studies often rely on general knowledge sources like Wikipedia to assess the models' abilities in solving common-sense problems. In this paper, we evaluated LLMs by RAG settings in a domain-specific context, college enrollment. We identified six required abilities for RAG models, including the ability in conversational RAG, analyzing structural information, faithfulness to external knowledge, denoising, solving time-sensitive problems, and understanding multi-document interactions. Each ability has an associated dataset with shared corpora to evaluate the RAG models' performance. We evaluated popular LLMs such as Llama, Baichuan, ChatGLM, and GPT models. Experimental results indicate that existing closed-book LLMs struggle with domain-specific questions, highlighting the need for RAG models to solve expert problems. Moreover, there is room for RAG models to improve their abilities in comprehending conversational history, analyzing structural information, denoising, processing multi-document interactions, and faithfulness in expert knowledge. We expect future studies could solve these problems better. 

\end{abstract}

\section{Introduction}
\label{sec:intro}
Recently, the emergence of large language models (LLMs) has revolutionized the way we access information. These LLMs are typically trained on vast amounts of web documents using the next token prediction task, which equips them with a wide range of world knowledge and advanced capabilities in understanding and generating natural language. However, despite these impressive attributes, they still face significant challenges, including hallucinations, difficulties in keeping up with real-time updates, etc~\cite{hanbench}. 

Retrieval-Augmented Generation (RAG), which involves retrieving external information from Information Retrieval (IR) systems to provide reliable knowledge, is a promising and widely adopted approach to overcome the above limitations. 
Furthermore, when deploying LLMs in practice, such as building question-answering systems for enterprises or some expert fields, it is more vital to provide domain-specific information for LLMs~\cite{zhang2024raft} since they are likely unequipped with this expert knowledge. 
For example, consulting firm financial statements or data aggregation in the investment industry are all widely used scenarios of RAG systems. Nevertheless, due to the problem of data privacy, these corpora cannot be incorporated into the training data of LLM, hence RAG systems are needed to plug these data into the LLMs in the form of external knowledge. 
Thus, evaluating the performance of RAG in domain-specific scenarios becomes imperative. However, existing studies~\cite{hanbench} predominantly rely on general knowledge sources, such as Wikipedia, as external knowledge bases to evaluate RAG models on dealing with commonsense or hot knowledge-intensive tasks~\cite{NQ,triviaqa,hotpotqa,KILT}. Such a method may not fully evaluate the ability of RAG models to solve domain-specific problems. 

\begin{figure*}
    \centering
    \includegraphics[width=0.95\linewidth]{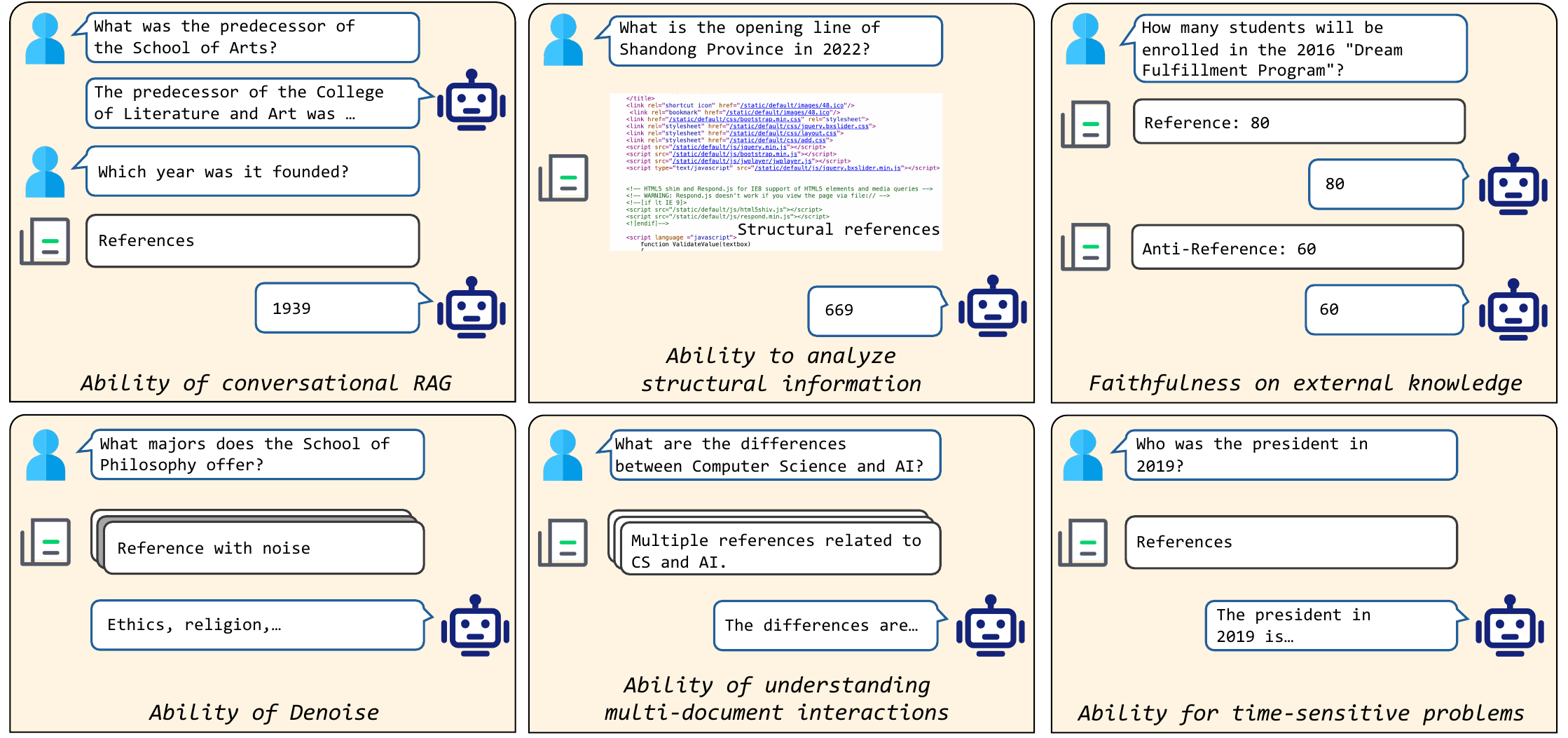}
    \caption{Important abilities for RAG models.}
    \label{fig:abilities}
\end{figure*}

Therefore, the use of domain-specific corpora and questions is essential to assess the ability of the LLM to effectively use external knowledge from these specific fields to solve expert problems.
In this paper, we identify six vital abilities to comprehensively evaluate RAG models, which are visualized in Figure~\ref{fig:abilities}, from three perspectives:

$\bullet$ \textbf{Understanding of user intents.} In traditional web information retrieval methods, such as search engines, understanding the actual user intents has always been a crucial step and studied in the literature~\cite{HTPS,PEPS,ExpliPS,HEXA,CIKM21_COCA,ASE,PnD,DIVPER,CAMI,CLPPS}. Nowadays, LLMs demonstrate remarkable abilities in various natural language processing tasks. However, comprehending user information needs and providing accurate responses is a more intricate task, especially in conversational scenarios that require clarifying the current user intents based on previous interactions. 
As a result, the \textit{conversation ability} is critical to building a user-friendly RAG system. 

$\bullet$ \textbf{Analysis of retrieved documents.} Apart from understanding user questions, the analysis of external documents plays a critical role in RAG systems. 
Considering that web pages not only contain massive textual knowledge but also intricate structures, such as HTML structures, which may also contain valuable information. It is also important for LLMs to \textit{comprehend the structural information} from the provided knowledge, hence providing accurate and reliable responses. 
Furthermore, the inherent difficulty for LLMs in acquiring in-domain knowledge underscores the importance of trusting external expert knowledge to bridge gaps in their own perception. In other words, when faced with in-domain problems, it is more reliable for LLMs to answer questions based on external expert knowledge rather than relying on their own knowledge, which may be limited and prone to hallucination. Thus, assessing the \textit{faithfulness of LLMs on external expert knowledge} is also an important task.

$\bullet$ \textbf{Interactions between intents and documents.} Given the provided external documents, LLMs must not only accurately comprehend the knowledge contained within them but also identify the relevant portions that contribute to solving the user's current problem. Typically, not all provided information is useful for solving problems, as there may be a significant amount of noise that potentially hinders the prediction of accurate results. Thus, assessing \textit{denoising ability} of RAG models is also critical. At the same time, this problem could be more distinct for time-sensitive questions, where the answers may change over time. Therefore, the RAG models's \textit{ability to solve time-sensitive problems} is another angle to evaluate their denoising abilities. Additionally, due to the complexity of user intents, answering some questions may require interactions between multiple documents and questions, highlighting the need for LLMs to effectively navigate and integrate information from various sources. As a result, we also propose to evaluate RAG models' \textit{ability to understand the interaction between multi-documents and complex questions}. 

Specifically, we constructed a comprehensive dataset that evaluates the above abilities of RAG models in a domain-specific scenario, namely \ours{}. The application scenario is the enrollment system of a university in China (with official permission). 
In addition to an extractive QA dataset that assesses basic QA ability, we further annotated the following sub-datasets, each targeting a specific ability, \ie, conversational QA, structural QA, faithful QA, time-sensitive QA, noisy QA, and multi-document QA. 
Concretely, the conversational QA dataset simulates complex and realistic scenarios where users interact with models through multiple turns to fulfill their information needs. 
The structural QA is designed to test the ability of LLMs to understand and infer answers from structured information of external knowledge, and the faithful QA evaluates the faithfulness of LLMs in handling external knowledge. 
The left three sub-datasets assess the capabilities of LLMs in handling the complex interaction between questions and documents. 
The noisy QA involves providing external knowledge with noisy information, challenging LLMs to filter out irrelevant or misleading content. The time-sensitive QA introduces time-sensitive questions, where the answers may vary at different timestamps. Lastly, the multi-document QA requires LLMs to integrate information from multiple external documents to provide satisfactory answers to complex questions.\footnote{Possibly, some sub-datasets may also indirectly evaluate LLMs' abilities from other perspectives. However, to decouple the assessment of each capability, we assign each sub-dataset to the category that best represents its primary focus.} 
In experiments, we evaluated seven popular LLMs, including Llama2-7B-chat, Llama2-13B-chat, Llama2-70B-chat, Baichuan2-7B-chat, Baichuan2-33B-32k, ChatGLM2-6B-32k, and GPT-3.5-turbo-1106. Generally, we find that 
(1) In domain-specific scenarios, most LLMs struggle to exactly answer the user questions without the aid of external knowledge. It highlights the importance of RAG models in such applications. 
(2) Leveraging HTML content is beneficial for LLMs to generate more accurate answers. However, the ability to comprehend and analyze structural information is not yet well-developed in all LLMs. Therefore, when deploying RAG models in practice, it is crucial to choose an LLM suitable for the specific application needs. 
(3) There is a large room for RAG models to improve their performance in complex scenarios involving various kinds of information sources. In conversational scenarios, RAG models need to accurately understand the user's intent based on historical information. In multi-doc QA, RAG models must comprehend the intricate relationships between multiple documents and questions. These challenges highlight the need for further investigation of high-quality RAG models.

\section{Related Work}
\label{sec:related_work}
\subsection{Retrieval-augmented Generation Models}
To alleviate the hallucination problem of language models, Retrieval-augmented generation (RAG) strategy~\cite{RAG_survey,llm_ir_survey} is proposed by providing external references to LMs to help them provide more accurate and factual answers. In particular, RAG approaches usually devise a retrieval model to collect relevant documents or passages according to user queries from the corpus. Then, these retrieved references are fed together with the user queries into the downstream language models to generate answers. 
Traditional approaches~\cite{realm,rag} in this area mainly focus on supporting language models with limited parameters such as BERT, BART, and T5. 
Recently, with the development of Large language models, more and more researchers have paid attention to improving the RAG performance based on large language models by considering the retrieval frequency~\cite{rag1,RALM,flare,retallm}, designing delicate CoTs~\cite{CON,IRCOT}, training the retrieval models and language models together~\cite{self-rag,radit}, comprising the retrieved references to fit the input length limit~\cite{compress1, compress2,compress3}. Research~\cite{rag_analy} has shown that by incorporating the retrieved documents, RAG models do respond with few mistakes.

\subsection{Evaluation of RAG}
Previous studies in the retrieval-augmented generation area mainly conduct experiments on open-domain QA datasets such NQ~\cite{NQ} and HotpotQA\cite{hotpotqa} using Wikipedia as the retrieval corpus. Though this general evaluation setting can somehow reflect the quality of answers generated by the RAG models, it fails to analyze the abilities of these models from different perspectives such as intent understanding and faithfulness to the references. Besides, since Wikipedia is widely used in the pre-training of language models and the information in the retrieved documents may have been learned by LLMs, it is questionable whether RAG models really utilize the references to answer questions instead of their intrinsic knowledge. Recently, Chen et al.~\cite{hanbench} alleviated this problem by proposing a specialized RAG benchmark to analyze the four disentangled fundamental abilities of different large language models. However, it still focuses on the open-domain QA, without considering the LLM's ability under in-domain situations. To thoroughly assess the abilities of RAG models, in this paper, we propose to leverage an in-domain document corpus collected from the enrollment websites of a Chinese university to evaluate the capabilities of RAG from multi-aspects.

\section{Evaluate Retrieval-Augmented Generation via In-domain Scenarios}\label{sec:method}
To avoid that the external knowledge has been studied well in pre-training or instruction tuning of LLMs, we focus on a domain-specific Chinese application scenario, \ie, college enrollment. This scenario primarily involves questions related to long-tail and domain-specific information such as admission introductions, admission policies, and details of schools or departments. Therefore, it is difficult for LLMs to rely solely on their internal knowledge to answer the user's questions. Instead, they need to heavily depend on external knowledge resources. To comprehensively evaluate the aforementioned capabilities of RAG models, we annotated seven sub-datasets, and the corresponding data construction process is demonstrated below. 

\subsection{Data Construction}
To acquire the document corpus for this scenario, we crawled web pages from the admission official website with official permission. We not only extracted their text contents but also reserved the original HTML contents to facilitate the construction of the structural QA dataset. Given the lengthy nature of web pages, we further split the text contents of each web page into passages using a sliding window of 256 length and 128 overlap. The numbers of web pages and passages are 1,686 and 14,406 respectively. Finally, we created two document corpora, a text corpus and an HTML corpus. The evaluation datasets are built by initially being generated from powerful generative models (ChatGPT or GPT-4), then being corrected manually.

$\bullet$ \textbf{Extractive QA Dataset.}
We first randomly sampled document passages from the text corpus. These passages were then incorporated into the prompt designed for ChatGPT, which generated question-answer (QA) pairs based on the provided passages. To ensure the in-domain nature of questions, we specifically instructed ChatGPT to generate questions that cannot be answered without providing external information. The selected passages can be directly considered as positive references for the corresponding questions in the dataset. 

$\bullet$ \textbf{Conversational QA Dataset.} 
To build question-answering conversations, we began by choosing documents with substantial content, primarily focusing on the introduction web pages of each school. We then utilized ChatGPT to generate domain-specific questions according to each passage within the selected documents. This process resulted in a collection of question-answer pairs associated with each document. To test the conversational intent understanding ability of RAG models, we simplified each question in the QA pair list (except the first one) by removing entities that duplicate the preceding questions. The revised QA list can be regarded as a vanilla conversation. Furthermore, we derived multiple conversation samples from a given vanilla one. Specifically, at the $t-$th step, the QAs from the previous $t-1$ steps were considered as historical conversations, and the current $t$-step QA was treated as the question and the golden answer. 

$\bullet$ \textbf{Structural QA Dataset.} 
To assess the understanding capabilities of RAG models on structural information (we focus on table structures in this paper), we first selected web pages containing table structures. Then, we offered the HTML contents of these web pages to ChatGPT and instructed it to generate QA pairs where the answers are derived from the table information. To accommodate the input length limits of ChatGPT, we preprocessed the HTML content by removing irrelevant elements, such as HTML comments, script tags, etc. For each QA pair, we provided both the HTML and corresponding pure text content of the positive document. This approach not only allows us to evaluate the models' ability to analyze and comprehend HTML structures but also to compare the effectiveness of using HTML structural information versus pure texts for solving problems.

$\bullet$ \textbf{Faithful QA Dataset.} 
To test the faithfulness of LLMs to domain-specific knowledge, we first provided document contents to GPT4 and prompt it to generate QA pairs that rely solely on external expert knowledge, rather than common-sense information. This step is similar to the process used in creating the extractive QA dataset. Furthermore, to ensure the generation quality, we manually filtered out the QA pairs that could be answered using knowledge contained within LLMs themselves. Finally, we modify the answer-related information in the positive references to build anti-references and corresponding anti-answers.
\begin{table}
    \small
    \centering
    \renewcommand{\arraystretch}{0.9}
    \caption{Overall results on the extractive, conversational, time-sensitive, and multi-doc datasets.}
    \begin{tabular}{llll}
    \toprule
    Dataset & Count & Avg. Q Len & Avg. A Len\\
    \midrule
    Extractive & 90 & 25.09 & 8.17 \\
    Conversational & 49 & 16.65 & 35.66 \\
    Structural & 94 & 35.48 & 6.07 \\
    Time-sensitive & 65 & 21.38 & 4.67 \\
    Multi-document & 48 & 35.90 & 86.69 \\
    Faithfulness & 49 & 27.29 & 12.85 /11.80 \\ 
    \bottomrule
    \end{tabular}
    \label{tab:statistic}
    \vspace{-5pt}
\end{table}

$\bullet$ \textbf{Noisy QA Dataset.} 
To evaluate LLMs' robustness to noisy information in provided references, we expanded upon the extractive QA dataset to create the noisy dataset. Concretely, for each piece of data, we randomly sampled several irrelevant passages from the text corpus to construct noisy information. During the test experiments, we varied the number of irrelevant passages selected and combined them with positive references to build external references with different noise ratios. 

$\bullet$ \textbf{Time-sensitive QA Dataset.} 
Since the dataset is static, it is difficult to evaluate the abilities of RAG in answering real-time questions. Inspired by~\cite{templama}, we focused on generating questions that have different answers at different timestamps. To indicate the timestamp of each question, we included a ``date'' attribute in each data sample. It is challenging for ChatGPT to automatically generate time-sensitive question-answer pairs that require rich prior domain knowledge. Therefore, we manually design possible questions and identify answerable document passages to build answers and positive references.

$\bullet$ \textbf{Multi-doc QA Dataset.} 
To address complex questions that can not be fully answered by simply extracting information at the entity level, it becomes necessary to aggregate information from multiple relevant documents. To build a dataset that evaluates such ability, we follow a specific approach. First, 
we identified a set of relevant documents that share similar topics or themes, such as introductions to relevant institutes or majors. These documents serve as the basis for generating the dataset. Next, we provide the text contents of these relevant documents to GPT4, which generates questions that require answers derived from multiple document contents. 

The statistical information of our datasets is demonstrated in Table~\ref{tab:statistic}. The noisy dataset is derived from the extractive dataset, its average lengths of queries and answers are the same as the extractive. There are two items of average answer length (Avg. A Len) of the faithful dataset where the first is the golden answer, the second is the anti-answer.\footnote{The anonymous link of our dataset is provided here: 
\href{https://github.com/ShootingWong/DomainRAG}{https://github.com/ShootingWong/DomainRAG}.}

\section{Experiment}
\subsection{Main settings}
We first conducted experiments using the following external knowledge settings,

(1) Close-book: No external domain-specific knowledge was provided to assess whether LLMs could solve these expert problems themselves.

(2) Golden reference: We provided human-annotated positive references for LLMs to explore the upper bounds of their abilities. 

(3) Retrieved reference: Simulating real-world applications of RAG models, we provided them with retrieved documents. We chose BM25~\cite{BM25} and BGE-base-zh-v1.5~\cite{bge_embedding} as two classical retrievers to represent sparse and dense retrieval. 

(4) Noisy reference. To test the robustness of LLMs on noisy external knowledge, we provided different levels of irrelevant references blended with golden references. We also investigated the impact of the position of golden references within the noisy references on RAG models' performances.

(5) Structural reference. In the experiments on the structural QA dataset, we provided two versions of golden references, \ie, HTML and pure texts, for LLMs to evaluate the abilities to analyze HTML structures and compare the effect of structural information versus pure texts for RAG tasks.

6) Anti-reference. In the faithful QA dataset, we provided both golden and anti-references for LLMs in the same question to compare the faithfulness of LLMs in utilizing expert external knowledge.

We selected six commonly used LLMs, including Llama2-7B-chat, Llama2-13B-chat, Llama2-70B-chat, Baichuan2-7B-chat, Baichuan2-33B-32k, ChatGLM2-6B-32k, and gpt-3.5-turbo-1106 to compare their abilities comprehensively. Note that we chose the Baichuan2-33B-32k version with a built-in general retrieval system to further assess the effectiveness of general knowledge sources in our domain-specific scenario.

\subsection{Evaluation Metrics}
To evaluate model performance, we chose four widely used metrics: Two versions of exact-match, where the one assesses whether the ground truth answers are contained by predictions (EM), the one assesses whether the predictions are strictly the same as the answers (EMS); F1 is used to evaluate models in the perspective of term-matching; Rouge-L and GPT-4 evaluation (GE) are used to assess the performance of long-form answers, \ie, conversational and multi-doc datasets.
For the GE metric, we prompt GPT to score whether the prediction is consistent with the answer from the three perspectives: factual consistency, redundancy, and deficiency. The predicted score should range from 0 to 5 and we normalized it to $[0,1]$.

\begin{table*}[!ht]
\tiny
    \centering
    \renewcommand{\arraystretch}{0.9}
    \caption{Overall results on the extractive, conversational, time-sensitive, and multi-doc datasets. The overall best result is indicated in bold, and the best result under each setting is identified with $^*$.}
    \begin{tabular}{@{}c@{}cllllllllllll@{}}
    \toprule
    
    \multirow{2}{*}{Settings} & \multirow{2}{*}{Models} & \multicolumn{4}{c}{Extractive} & \multicolumn{2}{c}{Conversational} & \multicolumn{4}{c}{Time-sensitive} & \multicolumn{2}{c}{Multi-doc} \\
    \cmidrule(lr){3-6}\cmidrule(lr){7-8} \cmidrule(lr){9-12} \cmidrule(lr){13-14}
    &  & EM & EMS & F1 & Rouge-L & Rouge-L & GE & EM & EMS & F1 & Rouge-L & Rouge-L & GE \\
    \midrule
    \multirow{7}{*}{\makecell{Close\\Book}} 
    & Llama2-7B-chat     & 0.1269 & 0.0000 & 0.1952 & 0.0863 & 0.1444 & 0.1429 & 0.1272 & 0.0000 & 0.1454 & 0.0706 & 0.2370 & 0.2750 \\
    & Llama2-13B-chat    & 0.1307 & 0.0000 & 0.2171 & 0.1018 & 0.1273 & 0.1878 & 0.1959 & 0.0000 & 0.1375 & 0.0411 & 0.2341 & 0.2624 \\
    & Llama2-70B-chat    & 0.1520 & 0.0000 & 0.2263 & 0.1096 & 0.1479 & 0.2122 & 0.1118 & 0.0000 & 0.1141 & 0.0426 & 0.2536 & 0.2542 \\
    & GPT-3.5-turbo-1106 & 0.1929 & 0.0111 & 0.3759 & 0.2102 & 0.2429$^*$ & 0.2245 & 0.0631 & 0.0154 & 0.2544 & 0.1177 & 0.2802 & 0.3292 \\
    & Baichuan2-7B       & 0.1548 & 0.0556 & 0.3531 & 0.1911 & 0.2108 & 0.2041 & 0.1118 & 0.0164 & 0.1620 & 0.0925 & 0.2397 & 0.2584 \\
    & ChatGLM2-6B-32K    & 0.1471 & 0.0000 & 0.1843 & 0.0781 & 0.1592 & 0.2082 & 0.1426 & 0.0154 & 0.1580 & 0.0880 & 0.2258 & 0.3208 \\
    & Baichuan2-33B-32k      & 0.2443$^*$ & 0.1333$^*$ & 0.4320$^*$ & 0.2828$^*$ & 0.1906 & 0.3143$^*$ & 0.2154$^*$ & 0.0769$^*$ & 0.2794$^*$ & 0.1722$^*$ & 0.2843$^*$ & 0.3334$^*$ \\
    \midrule
    \multirow{7}{*}{\makecell{Golden\\Reference}} 
    & Llama2-7B-chat     & 0.7986 & 0.0111 & 0.3948 & 0.3503 & 0.4460 & 0.5388 & 0.7405 & 0.0154 & 0.4166 & 0.3701 & 0.2626 & 0.3750 \\
    & Llama2-13B-chat    & 0.8106 & 0.0000 & 0.5322 & 0.4745 & 0.5004 & 0.6858 & 0.7867 & 0.0000 & 0.5161 & 0.4497 & 0.2950 & 0.4666 \\
    & Llama2-70B-chat    & 0.8880 & 0.2556 & 0.6612 & 0.6219 & 0.5762 & 0.7918 & 0.8846 & 0.4923 & 0.7364 & 0.7056 & 0.3179 & 0.4958$^*$ \\
    & GPT-3.5-turbo-1106 & \textbf{0.9233}$^*$ & 0.3667 & 0.8213 & 0.8065 & 0.5963 & 0.7633 & 0.8785 & 0.6308 & 0.8811 & 0.8711 & 0.3901 & 0.4917 \\
    & Baichuan2-7B       & 0.7794 & 0.4583 & 0.7718 & 0.7044 & 0.5231 & 0.7061 & 0.7923 & 0.5846 & 0.7923 & 0.7450 & 0.3392 & 0.4375 \\
    & ChatGLM2-6B-32K    & 0.8503 & 0.0556 & 0.4845 & 0.4528 & 0.5797 & 0.7674 & 0.7123 & 0.2000 & 0.4175 & 0.3915 & 0.3357 & 0.4124 \\
    & Baichuan2-33B-32k      & 0.8667 & \textbf{0.5778}$^*$ & \textbf{0.8885}$^*$ & \textbf{0.8674}$^*$ & \textbf{0.6632}$^*$ & \textbf{0.8326}$^*$ & \textbf{0.9154}$^*$ & \textbf{0.7846}$^*$ & 0\textbf{.9503}$^*$ & \textbf{0.9459}$^*$ & \textbf{0.3936}$^*$ & \textbf{0.4958}$^*$ \\
    \midrule
    \multirow{7}{*}{\makecell{BM25\\TOP1}}
    & Llama2-7B-chat     & 0.6638 & 0.0111 & 0.3647 & 0.2942 & 0.2764 & 0.1429 & 0.4246 & 0.0000 & 0.2618 & 0.2156 & 0.2371 & 0.3042 \\
    & Llama2-13B-chat    & 0.6988 & 0.0000 & 0.4621 & 0.3847 & 0.3125 & 0.3674 & 0.4087 & 0.0308 & 0.2911 & 0.2488 & 0.2587 & 0.2708 \\
    & Llama2-70B-chat    & 0.7184 & 0.2111 & 0.5778 & 0.5029 & 0.3467 & 0.4654$^*$ & 0.4497 & 0.2000 & 0.4422 & 0.3721 & 0.3039 & 0.3458$^*$ \\
    & GPT-3.5-turbo-1106 & 0.7749$^*$ & 0.3222 & 0.7222 & 0.6588 & 0.3798 & 0.7552 & 0.4800$^*$ & 0.3385 & 0.5273$^*$ & 0.4812 & 0.2647 & 0.2958 \\
    & Baichuan2-7B       & 0.6562 & 0.4028 & 0.6916 & 0.5980 & 0.3271 & 0.4000 & 0.4215 & 0.3472 & 0.5111 & 0.4456 & 0.2819 & 0.2500 \\
    & ChatGLM2-6B-32K    & 0.7029 & 0.0111 & 0.3954 & 0.3384 & 0.3261 & 0.4612 & 0.4369 & 0.0615 & 0.2498 & 0.2057 & 0.2683 & 0.3208 \\
    & Baichuan2-33B-32k      & 0.7267 & 0.5111$^*$ & 0.7717$^*$ & 0.7045$^*$ & 0.4092$^*$ & 0.4000 & 0.4579 & 0.4236$^*$ & 0.5272 & 0.5140$^*$ & 0.3074$^*$ & 0.2958 \\
    \midrule
    \multirow{7}{*}{\makecell{BM25\\TOP3}}
    & Llama2-7B-chat     & 0.7188 & 0.0000 & 0.2881 & 0.2373 & 0.2419 & 0.2734 & 0.5390 & 0.0000 & 0.1794 & 0.1463 & 0.2219 & 0.3376 \\
    & Llama2-13B-chat    & 0.7548 & 0.0000 & 0.3203 & 0.2616 & 0.2652 & 0.4734 & 0.5795 & 0.0000 & 0.2146 & 0.1769 & 0.2454 & 0.3584 \\
    & Llama2-70B-chat    & 0.7298 & 0.0111 & 0.4038 & 0.3409 & 0.3074 & 0.4490 & 0.5672 & 0.0462 & 0.3450 & 0.2853 & 0.2651 & 0.3792$^*$ \\
    & GPT-3.5-turbo-1106 & 0.7835$^*$ & 0.3222 & 0.7463 & 0.6951 & 0.4200 & 0.5306 & 0.6400 & 0.5077 & 0.6805 & 0.6397 & 0.3419$^*$ & 0.3542 \\
    & Baichuan2-7B       & 0.7145 & 0.4333 & 0.6814 & 0.6106 & 0.3281 & 0.5756$^*$ & 0.5067 & 0.3646 & 0.5412 & 0.4824 & 0.2918 & 0.3876 \\
    & ChatGLM2-6B-32K    & 0.7245 & 0.0333 & 0.4868 & 0.4352 & 0.3758 & 0.4530 & 0.4959 & 0.1231 & 0.3575 & 0.3110 & 0.2806 & 0.3708 \\
    & Baichuan2-33B-32k      & 0.7767 & 0.5593$^*$ & 0.8014$^*$ & 0.7568$^*$ & 0.4352$^*$ & 0.5756$^*$ & 0.6841$^*$ & 0.6369$^*$ & 0.7284$^*$ & 0.7226$^*$ & 0.3406 & 0.3584 \\
    \midrule
    \multirow{7}{*}{\makecell{Dense\\TOP1}}
    & Llama2-7B-chat     & 0.5048 & 0.0000 & 0.3160 & 0.2210 & 0.2182 & 0.2898 & 0.3564 & 0.0000 & 0.2093 & 0.1383 & 0.2489 & 0.3458 \\
    & Llama2-13B-chat    & 0.5651 & 0.0111 & 0.4210 & 0.3088 & 0.2860 & 0.3020 & 0.3867 & 0.0000 & 0.3062 & 0.2351 & 0.2728 & 0.3584 \\
    & Llama2-70B-chat    & 0.5807$^*$ & 0.1444 & 0.5086 & 0.3967 & 0.3010 & 0.3674 & 0.4026$^*$ & 0.1846 & 0.3706 & 0.2978 & 0.2896 & 0.3792$^*$ \\
    & GPT-3.5-turbo-1106 & 0.5467 & 0.2235 & 0.5915 & 0.4831 & 0.3180 & 0.3632 & 0.3979 & 0.2769 & 0.4874$^*$ & 0.4031$^*$ & 0.3327$^*$ & 0.3792$^*$\\
    & Baichuan2-7B       & 0.4499 & 0.2444 & 0.5220 & 0.4129 & 0.2771 & 0.3633 & 0.2733 & 0.2579 & 0.3539 & 0.3129 & 0.2790 & 0.3708 \\
    & ChatGLM2-6B-32K    & 0.5204 & 0.0222 & 0.3310 & 0.2406 & 0.2666 & 0.3062 & 0.3410 & 0.0154 & 0.1857 & 0.1242 & 0.2666 & 0.3624 \\
    & Baichuan2-33B-32k     & 0.5573 & 0.3333$^*$ & 0.6467$^*$ & 0.5192$^*$ & 0.3772$^*$ & 0.4858$^*$ & 0.3831 & 0.3215$^*$ & 0.4631 & 0.3943 & 0.2970 & 0.3208 \\
    \midrule
    \multirow{7}{*}{\makecell{Dense\\TOP3}}
    & Llama2-7B-chat     & 0.5630 & 0.0000 & 0.2601 & 0.2022 & 0.2357 & 0.3388 & 0.4328 & 0.0000 & 0.1436 & 0.1112 & 0.2221 & 0.3458 \\
    & Llama2-13B-chat    & 0.6111 & 0.0000 & 0.2763 & 0.2178 & 0.2678 & 0.4286 & 0.4077 & 0.0000 & 0.1462 & 0.1197 & 0.2621 & 0.4458 \\
    & Llama2-70B-chat    & 0.6921 & 0.0111 & 0.3920 & 0.3265 & 0.2775 & 0.4490 & 0.4359 & 0.0000 & 0.3480 & 0.2749 & 0.2549 & 0.4250 \\
    & GPT-3.5-turbo-1106 & 0.6930$^*$ & 0.3111 & 0.7096 & 0.6364 & 0.4028 & 0.4898 & 0.4021 & 0.2769 & 0.4814 & 0.4112 & 0.3580$^*$ & 0.4334 \\
    & Baichuan2-7B       & 0.5917 & 0.3000 & 0.5805 & 0.4938 & 0.3327 & 0.4612 & 0.3636 & 0.2359 & 0.4255 & 0.3635 & 0.2916 & 0.4876$^*$ \\
    & ChatGLM2-6B-32K    & 0.6635 & 0.0333 & 0.4530 & 0.3922 & 0.3390 & 0.4694 & 0.4308 & 0.1385 & 0.3295 & 0.2868 & 0.2837 & 0.3500 \\
    & Baichuan2-33B-32k      & 0.6769 & 0.4222$^*$ & 0.7299$^*$ & 0.6533$^*$ & 0.4696$^*$ & 0.5755$^*$ & 0.4841$^*$ & 0.3692$^*$ & 0.5633$^*$ & 0.4839$^*$ & 0.3384 & 0.4000 \\
    \bottomrule
    \end{tabular}
    \label{tab:overall}
    \vspace{-5pt}
\end{table*}

\subsection{Overall Experimental Results}
The overall experimental results of the first three settings are shown in Table~\ref{tab:overall}, and we analyze the following conclusions.

(1) In domain-specific scenarios, the knowledge contained within LLMs themselves may hard to tackle the user's expert problems. 
The experimental results in the "Close Book" block confirm the poor performance of LLMs when faced with in-domain questions that go beyond their internal knowledge
Additionally, 
the retrieval settings in the last four blocks demonstrate that external expert knowledge can provide more reliable information for LLMs in expert scenarios. 
Even when equipped with a built-in retrieval system like Baichuan2-33B-32k, the close-book results are significantly inferior to those obtained from retrieval settings. This finding reinforces the importance of domain-specific corpora over general knowledge sources.

(2) The BM25 retriever shows better generalization than the dense retriever. Interestingly, when using BM25, the results are generally better than the dense retriever. 
One possible explanation is that our application scenarios are long-tail, meaning they contain specialized knowledge that may not have been adequately covered during the pre-training of the dense retriever.
However, BM25 is a spare retrieve with strong generalization ability. Therefore, the BM25 is also a good choice for RAG models in domain-specific application scenarios, especially if applications are more cost-sensitive.

(3) Dealing with long-form QA problems remains challenging for RAG models and warrants further investigation in the future. We notice that for conversational and multi-doc datasets, the improvement of retrieval-augmented results is not as significant as in other datasets. Especially for the multi-doc dataset, its results of providing golden references are still limited. These phenomena imply that accurately understanding the user's current intents in conversational scenarios requires RAG models to possess strong abilities in analyzing complex relationships among historical information, the current query, and external references. Moreover, analyzing the interactions between multiple external documents is also a critical ability for RAG models and needs to be investigated further. 
\begin{figure}
    \centering
    \includegraphics[width=1\linewidth]{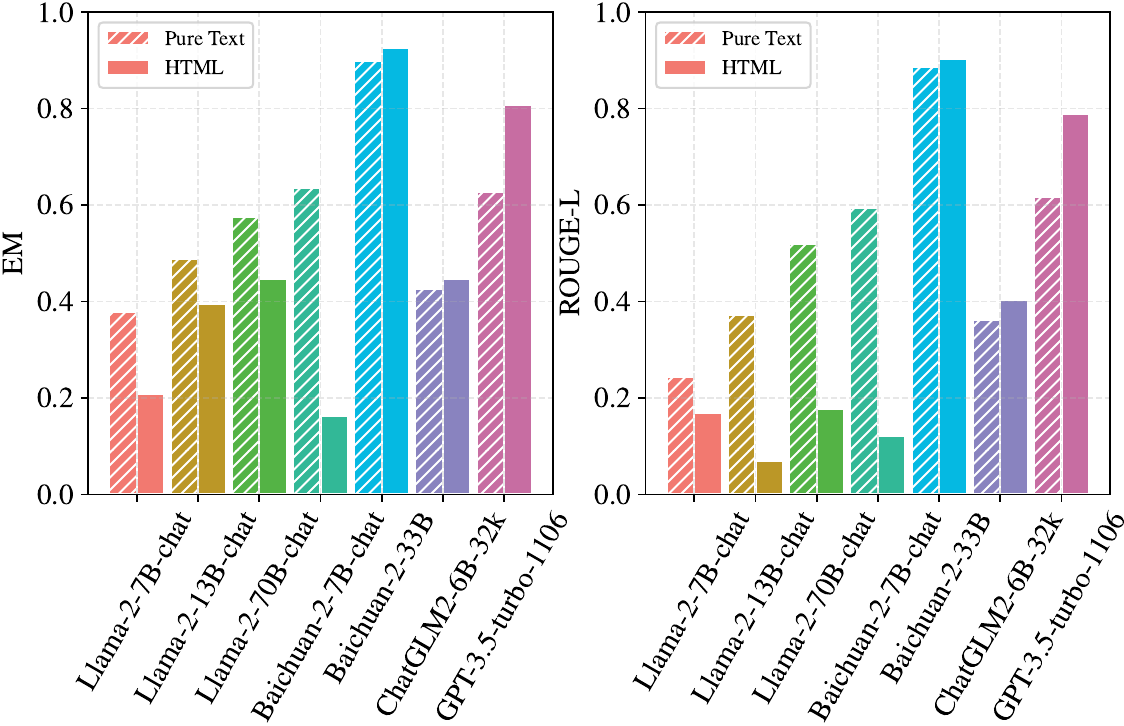}
    \caption{The experiments on the structural QA dataset.}
    \label{fig:struct}
\end{figure}
\begin{figure}
    \centering
    \includegraphics[width=1\linewidth]{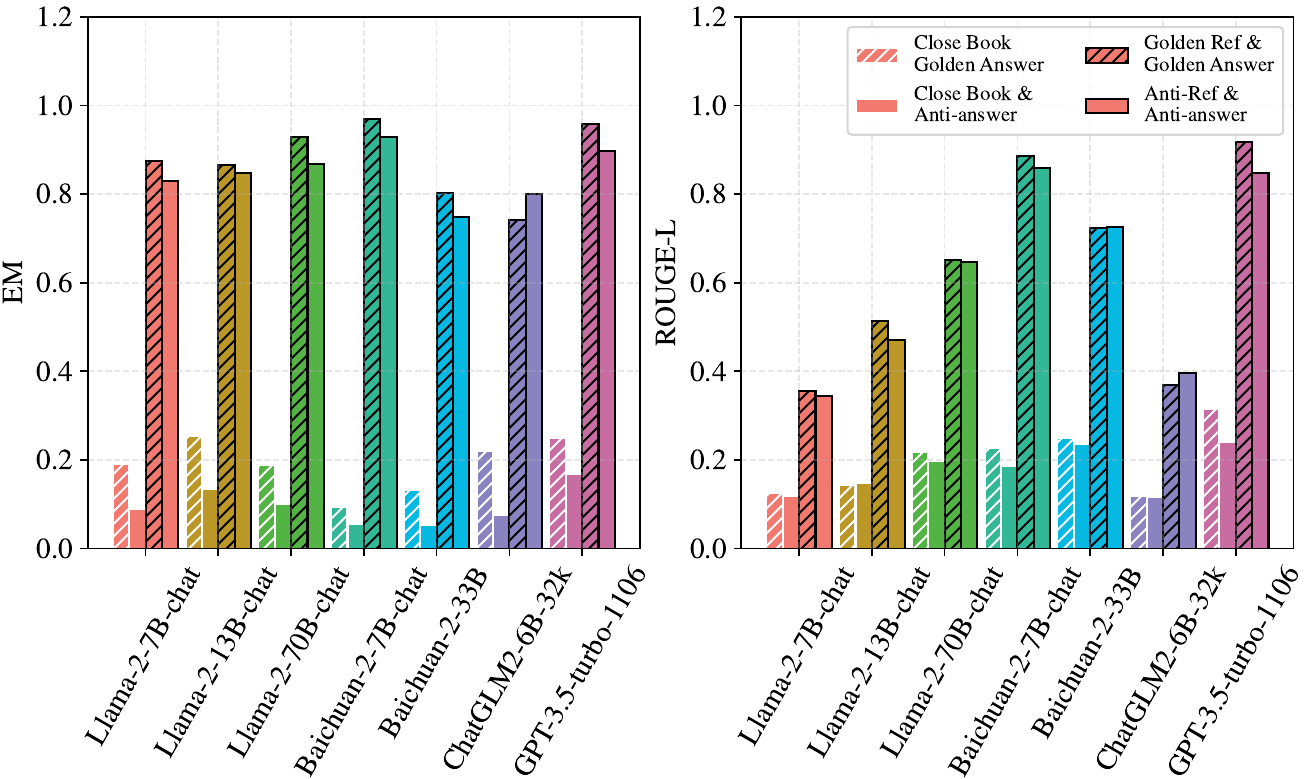}
    \caption{The experiments on the faithful QA dataset.}
    \label{fig:counter}
    \vspace{-5pt}
\end{figure}
\subsection{Experiments on Structural Dataset}
To evaluate the effectiveness of structural information for RAG models and analyze their abilities to comprehend knowledge in HTML format, we conducted the corresponding experiments on our structural QA dataset. It is worth noting that the whole HTML content of a web page is redundant and may contain some useless information about the web layout. Therefore, we proactively filtered out the information irrelevant to the valuable content of web pages. Nevertheless, the processed contents still exceed the maximum length of some LLMs, \eg, Llama. For simplicity, we directly truncated the provided information for LLMs that cannot handle lengthy texts. We expect that there are more elaborate techniques to tackle this problem. We provided two versions of web page content: one is pure text and the other one is HTML content for comparing the performance of RAG models on the different formats of external knowledge. The experimental results are presented in Figure~\ref{fig:struct}.

Obviously, for Llama models and Baichuan2-7B, the predicted results from pure texts are more accurate than ones from HTML content. However, for Baichuan2-33B-32k, ChatGLM2-6B-32k, and GPT-3.5-turbo-1106, providing HTML content leads to better performance than providing pure texts. According to this phenomenon, we can draw the following conclusions:
(1) Structural data contains valuable information beyond pure texts. Strong LLMs, such as GPT-3.5-turbo-1106, demonstrate better results when using HTML-format external knowledge, suggesting that the structural information of web pages complements textual content and helps LLMs understand web content and address user queries effectively.
(2) Baichuan2-33B-32k, ChatGLM2-6B-32k, and GPT-3.5-turbo-1106 have stronger abilities in understanding and analyzing HTML contents than Llama-family models and Baichuan2-7B. The reason may be that these models have been pre-trained on data in the HTML format, which enables them to better comprehend the corresponding information. In the future, with more diverse formats of external knowledge, such ability is more and more important for LLMs to provide better experiences for users.

\subsection{Robustness of LLMs on Noisy References}
To assess the robustness of LLMs on noised references, we mixed the positive references with different amounts of noisy references, including 4, 9, 14, 19, and 24.
Additionally, the position of the positive reference was varied, \ie the first, the middle, and the last positions, to assess the impact of the reference order on RAG models. 
The experiments were performed on LLMs capable of processing long texts, \ie Baichuan2-33B and GPT-3.5-turbo-1106, considering the potential issue of overlength when dealing with extensive external knowledge.

\begin{figure}[!ht] 
    \centering 
    \includegraphics[width=1\linewidth]{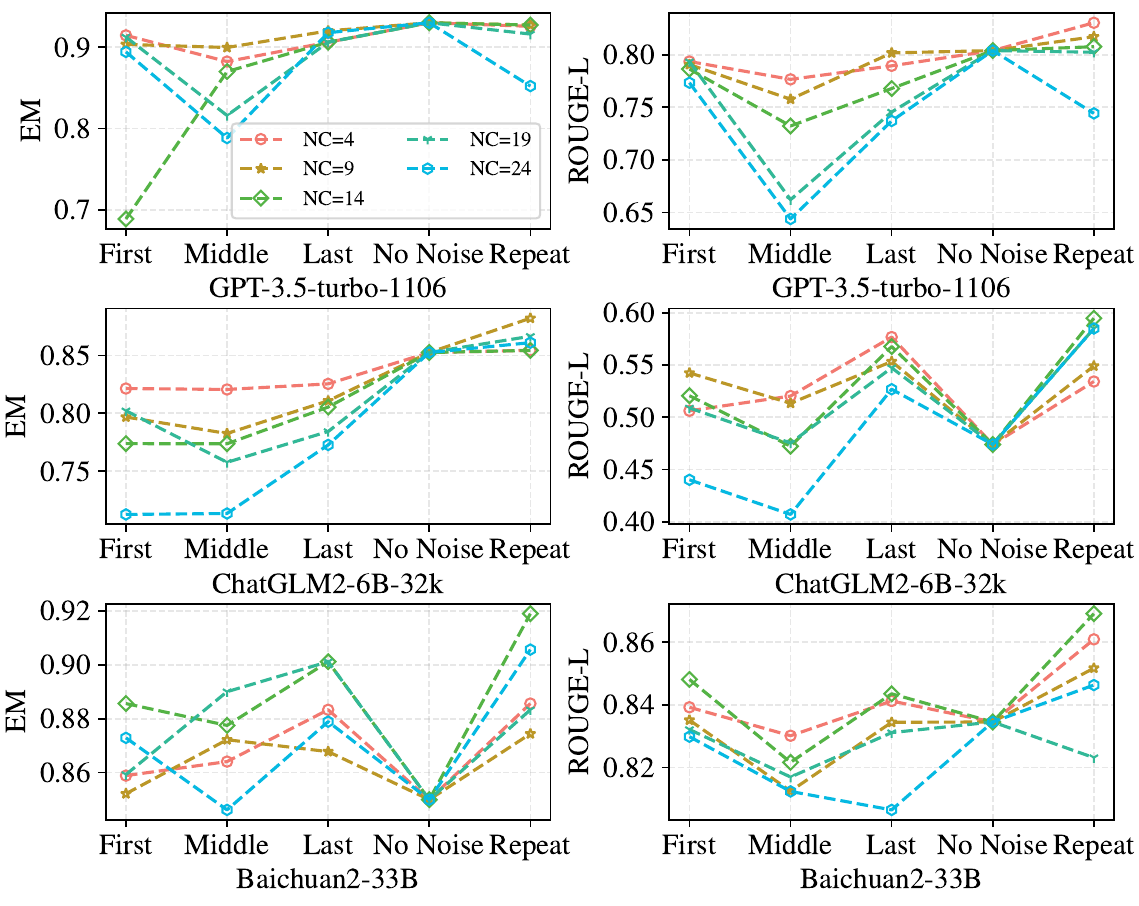} 
    \caption{Experiments in different noise ratio settings.} 
    \label{fig:noisy}
    \vspace{-5pt}
\end{figure}

The results in Figure~\ref{fig:noisy} indicate that both different positions of golden references and amounts of noise have a significant influence on the performance of RAG models. There are some interesting findings: (1) Lost in the middle is a common phenomenon. Placing positive references in the middle position of noisy references often leads to a significant decline in model performance. This phenomenon has also been indicated in recent studies~\cite{LostinMiddle}, highlighting the importance of not only the quality of the provided knowledge but also its order. (2) More noise often leads to worse results. It is evident that the lines with high $NC$-values are generally below the line with low $NC$-values (``NC'' is short for ``Noise Count''). It is reasonable since excessive irrelevant knowledge may disturb the LLMs' cognition, thus negatively affecting the generated results. Therefore, a high-quality IR model is also critical for RAG tasks. (3) Noise is not always bad. The results of ``No Noise'' are not always the best compared to those obtained from noisy references. The reason may be that 
compared to the ``No Noise'' setting where only one document is provided, the noisy references contain $NC+1$ external documents.
The increased amount of provided knowledge may emphasize the confidence of LLMs in external knowledge, making them more inclined to rely on it when solving problems. To verify this assumption, we conducted an experiment, where the golden references were repeated to match the number of noisy references. This experiment partially supports this assumption as the repeated references outperformed all other settings in most situations. This observation provides some insights for future studies of RAG that repeating provided references may be beneficial for motivating LLMs to provide better results.

\subsection{Faithfulness of LLMs in External References}
To assess the faithfulness of LLMs in external knowledge in out-of-domain applications, we provided the anti-references for LLMs to test whether they could generate anti-answers for these expert questions according to the external information. We compare the results with two different settings, in one we provided golden references and tested the performance of generating golden answers, other one is the close book setting. The comparison results are demonstrated in Figure~\ref{fig:counter}.
We found that in the close book setting, LLMs significantly underperform the settings with external knowledge, further confirming the importance of external knowledge for this scenario. Additionally, whether or not external knowledge is provided, LLMs often tend to generate golden answers instead of anti-answers. This suggests that (1) LLMs still contain a certain of prior knowledge about the in-domain information. (2) There is some room for LLMs to improve their faithfulness in the right external knowledge. 
It is reasonable that LLMs could answer some questions according to their own knowledge, hence may impact their confidence in external information. Nevertheless, their own knowledge is static and may be out-of-date while knowledge in some in-domain scenarios, such as enrollment plans, will quickly change over time. These situations put forward a strong requirement for LLMs to distinguish and trust external knowledge correctly.

\section{Conclusion}
We built a comprehensive dataset, \ours{}, to assess some crucial abilities of RAG models in a domain-specific scenario, college enrollment. We crawled the corresponding webpages from the website and two types of corpora, HTML corpus and pure text corpus were built. Then, we created corresponding sub-datasets to assess the following abilities, \ie conversational RAG, structural information analysis, faithfulness to external knowledge, denoising, solving time-sensitive problems, and understanding of multi-document interactions. Our experiments confirm the role of RAG models in domain-specific scenarios where LLMs cannot solve expert questions well. Furthermore, 
RAG models still have room for improvement in comprehending users' conversational history, analyzing structural knowledge, denoising references, managing multi-document interactions, and preserving fidelity to expert knowledge. We expect future research to make advancements in addressing these challenges more effectively.



\newpage

\section{Limitations}
In this work, we identified six critical capabilities of RAG models and developed a comprehensive dataset, namely \ours{}, to evaluate these capabilities in a domain-specific application scenario. We acknowledge the following limitations of our current study that present opportunities for future investigations.

First, though we chose several popular LLMs to assess their abilities in leveraging external knowledge to solve domain-specific questions, there exists some more sophisticated frameworks designed for enhancing the performance of RAG systems. Due to the complexity and diversity of implementation processes, we did not include them in our current research and evaluate their performances. Secondly, the application scenario is single. While we selected a in-domain and long-tail application scenario, its unicity may also introduce some biases to experimental results. In the future, it is valuable to explore more model structures and application scenarios to evaluate the capabilities of RAG systems more comprehensively and reliably.

\bibliography{custom}




\end{document}